%% file: paper.tex
\documentclass[10pt,a4paper,logo]{paper}

\usepackage[all]{hypcap}

\usepackage[authoryear, round]{natbib}

\usepackage{nicefrac}       
\usepackage{multirow}
\usepackage{cleveref}
\usepackage{dsfont}
\usepackage{makecell}
\usepackage{subcaption}
\usepackage{bbm}
\usepackage{wrapfig}
\usepackage{mathtools,xparse}
\usepackage{array}
\usepackage{scalerel}
\usepackage{tablefootnote}
\usepackage{tocloft}
\usepackage{adjustbox}
\usepackage{setspace}
\usepackage[most,skins,theorems]{tcolorbox}

\usepackage{placeins}
\usepackage[table]{xcolor}
\usepackage{soul}

\newtcolorbox{examplebox}[2][]{
  enhanced, unbreakable,
  colback=white, colframe=boxframe, boxrule=0.5pt, arc=3pt,
  colbacktitle=boxhead, coltitle=black, titlerule=0.5pt,
  fonttitle=\small\bfseries, title={#2},
  left=8pt, right=8pt, top=6pt, bottom=6pt, boxsep=0pt,
  lefttitle=8pt, toptitle=4pt, bottomtitle=4pt, #1
}

\newboolean{showsection}
\setboolean{showsection}{true}
\makeatletter
\@namedef{ver@everyshi.sty}{}
\makeatother

\definecolor{custom_green}{rgb}{0.0, 0.5, 0.0}
\definecolor{custom_red}{rgb}{1.0, 0.01, 0.24}
\definecolor{custom_blue}{HTML}{C9DAF7}
\definecolor{custom_purple}{HTML}{D9D1E9}
\definecolor{title_blue}{HTML}{204899}
\definecolor{cite_blue}{HTML}{044dc1}
\definecolor{cite_purple}{HTML}{7406a7}

\hypersetup{
    colorlinks = true,
    citecolor = {cite_blue},
    linkcolor = {cite_purple},
    urlcolor = {cite_purple},
}

\input{macro}

\input{math_commands.tex}

\makeatletter
\def\mathcolor#1#{\@mathcolor{#1}}
\def\@mathcolor#1#2#3{%
  \protect\leavevmode
  \begingroup
    \color#1{#2}#3%
  \endgroup
}
\makeatother

\Crefformat{equation}{#2Eq.\;(#1)#3}
\Crefformat{figure}{#2Figure #1#3}
\Crefformat{assumption}{#2Assumption #1#3}
\Crefname{assumption}{Assumption}{Assumptions}

\usepackage{crossreftools}
\makeatletter
\renewcommand\footnoterule{%
  \kern 15\p@
  \hrule \@width 2in \kern 2.6\p@
  \vspace{4pt}
}
\makeatother

\title{\tulip~TULIP: Targeted LLM Unlearning at Layers Identified Per-input}

\reportnumber{}

\author[1]{Yejin Kim}
\author[2]{William F. Shen}
\author[1]{Seokwon Jung}
\author[3]{Daeun Park}
\author[1]{Seong Joon Oh}

\affil[1]{KAIST}
\affil[2]{University of Cambridge}
\affil[3]{Sookmyung Women's University}

\correspondingauthor{Correspondence to: \email{yejin.c.kim@kaist.ac.kr}.}

\input{sections/1_abstract}

\begin{document}

\maketitle

\input{sections/2_introduction}
\input{sections/3_related_work}
\input{sections/4_method}
\input{sections/5_experimental_result}
\input{sections/6_analysis}
\input{sections/7_conclusion}

\bibliographystyle{plainnat}
\bibliography{paper}

\clearpage
\appendix{\input{sections/8_appendix}}

\end{document}

%% file: macro.tex
\definecolor{blanchedalmond}{rgb}{1.0, 0.92, 0.8}
\definecolor{carmine}{rgb}{0.59, 0.0, 0.09}
\definecolor{lightblue}{rgb}{0.22,0.45,0.70}%

\renewcommand{\mathbf}{\boldsymbol}

\makeatletter
\def\Ddots{\mathinner{\mkern1mu\raise\p@
\vbox{\kern7\p@\hbox{.}}\mkern2mu
\raise4\p@\hbox{.}\mkern2mu\raise7\p@\hbox{.}\mkern1mu}}
\makeatother

\numberwithin{equation}{section}

\definecolor{amaranth}{rgb}{0.9, 0.17, 0.31}
\definecolor{antiquebrass}{rgb}{0.8, 0.58, 0.46}
\definecolor{antiquefuchsia}{rgb}{0.57, 0.36, 0.51}
\definecolor{chromeyellow}{rgb}{0.31, 0.47, 0.26}

\newcommand{\1}{\mathds 1}

\newcommand{\tulip}{\texorpdfstring{\raisebox{-0.12em}{\includegraphics[height=0.9em]{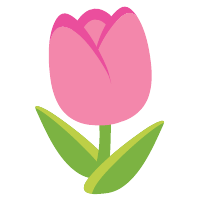}}}{}}

\definecolor{boxframe}{gray}{0.78}
\definecolor{boxhead}{gray}{0.96}
\definecolor{redhl}{RGB}{252,212,200}
\definecolor{greenhl}{RGB}{214,238,214}
\definecolor{yellowhl}{RGB}{255,241,168}
\definecolor{grayhl}{gray}{0.9}

\newcommand{\grayhl}[1]{{\sethlcolor{grayhl}\hl{#1}}}
\newcommand{\mlabel}[1]{\textbf{#1}}
\newcommand{\redhl}[1]{{\sethlcolor{redhl}\hl{#1}}}
\newcommand{\greenhl}[1]{{\sethlcolor{greenhl}\hl{#1}}}
\newcommand{\yellowhl}[1]{{\sethlcolor{yellowhl}\hl{#1}}}
\newcommand{\caphl}[2]{{\setlength{\fboxsep}{1pt}\colorbox{#1}{\strut #2}}}

%% file: math_commands.tex
\usepackage{amsmath,amsfonts,bm}

\def\eqref#1{equation~\ref{#1}}

\def\1{\bm{1}}

\DeclareMathAlphabet{\mathsfit}{\encodingdefault}{\sfdefault}{m}{sl}
\SetMathAlphabet{\mathsfit}{bold}{\encodingdefault}{\sfdefault}{bx}{n}



%% file: sections/1_abstract.tex
\begin{abstract}
Representation-level unlearning modifies the intermediate hidden states of LLMs. Although knowledge is distributed across layers, existing methods operate at a \textit{single fixed layer} for the entire forget set. We ask whether such a fixed layer is sufficient for unlearning. To answer this, we design a hijacking experiment that grafts hidden states of the target model into an oracle model. The oracle cannot produce the forget answer on its own, yet it produces the answer from the grafted state. This indicates the answer is already \emph{formed} at an intermediate layer and \emph{read out} afterward. Moreover, the layer where formation ends varies widely across inputs. Motivated by these findings, we propose \textbf{Targeted Unlearning at Layers Identified Per-input (TULIP)}. For each input, TULIP uses the logit lens to locate the formation--readout boundary and removes the hidden state's alignment with the forget answer's unembedding vector at that layer. TULIP outperforms output- and representation-level baselines on TOFU, PISTOL, and WMDP across Llama, Qwen, and Zephyr models. It also remains robust to paraphrase and quantization attacks. Beyond standalone use, its per-input layer selection serves as a plug-and-play component that further improves existing methods.
\end{abstract}

%% file: sections/2_introduction.tex
\section{Introduction}
\label{1_introduction}

Machine unlearning aims to modify a trained model so that it behaves as if a designated subset of its training data had never been used \citep{pmlr-v119-guo20c, bourtoule2021machine, sekhari2021remember}. Early approaches pursue this goal at the output level, penalizing the model's responses associated with the forget data \citep{jang-etal-2023-knowledge, zhang2024negative}. However, these approaches can change what the model says without removing what it knows, leaving the underlying information recoverable \citep{lucki2024adversarial, wang2025invariance, xu2026unlearning}. More recent work, which we refer to as \textit{representation-level unlearning}, directly modifies the intermediate representations instead, where this knowledge is encoded \citep{Li2024wmdp, huu2025adarmu, shen2025llm}. In doing so, they aim to remove the knowledge itself rather than merely suppress its expression.

Representation-level unlearning methods typically intervene at a \emph{single fixed layer} for the entire forget set. RMU applies its intervention at one layer chosen by hyperparameter search \citep{Li2024wmdp}. Adaptive RMU follows the same layer selection but adapts the steering coefficient to each forget input's representation norm, making RMU effective at later layers as well \citep{huu2025adarmu}. LUNAR selects the single layer that best elicits refusal-style responses \citep{shen2025llm}. Whatever the selection criterion, the chosen layer is applied uniformly to every forget instance.

Is a single intervention layer sufficient to unlearn the entire forget set? To examine this, we design a \emph{hijacking} experiment that grafts the target model's intermediate hidden state into the oracle at the same layer. With this grafted state, the oracle produces the target token that it cannot produce on its own. This indicates that an LLM's input-to-output mapping can be decomposed into two stages. In \emph{formation}, the target token is encoded into the hidden state, and in \emph{readout}, the remaining layers decode it. Since even the oracle can perform readout, unlearning should disrupt formation while preserving the readout path. Moreover, the layer at which formation ends and readout begins varies widely with the input context (Figure~\ref{fig:hijack}(b)). A single intervention layer is therefore insufficient, and the intervention layer should be chosen per input.

Motivated by these observations, we propose \textbf{T}argeted \textbf{U}nlearning at \textbf{L}ayers \textbf{I}dentified \textbf{P}er-input (TULIP), which unlearns each forget instance at its own formation--readout boundary. TULIP estimates this boundary for each input with the logit lens, as the earliest layer from which the target remains the top-1 token. At this layer, TULIP removes the alignment between the hidden state and the target's unembedding vector, while a retain term preserves hidden states on the retain set.

We evaluate TULIP on TOFU \citep{maini2024tofu}, PISTOL \citep{qiu2025pistol}, and WMDP \citep{Li2024wmdp}. The backbones span the Llama \citep{touvron2023llama2, grattafiori2024llama3}, Qwen \citep{qwen2025qwen25technicalreport, yang2025qwen3technicalreport}, and Zephyr \citep{tunstall2024zephyr} families and range from 1B to 8B parameters. Our method outperforms the baselines across benchmarks and model sizes while preserving model utility. It also remains robust to paraphrase \citep{maini2024tofu, wang-etal-2026-erasing} and quantization \citep{zhang2025catastrophic} attacks. Further analysis shows that dynamic layer selection drives these gains and also works as a plug-and-play component for existing unlearning methods. We summarize our contributions as follows:
\begin{itemize}
  \item We redefine what unlearning should remove. Decomposing how a model produces a forget target into \emph{formation} and \emph{readout}, we argue that unlearning should remove formation while preserving readout. A hijacking experiment that grafts the target model's hidden states into an oracle supports this view, showing that the target is already formed at an intermediate layer and that even the oracle can read it out from there. The experiment further reveals that the depth of this layer varies widely across inputs.
  \item We propose TULIP\tulip, which estimates the formation--readout boundary of each input with the logit lens and removes the alignment between the hidden state and the target at that layer, thereby unlearning each instance where its target is formed.
  \item We demonstrate that TULIP outperforms output-level and representation-level unlearning methods across TOFU, PISTOL, and WMDP, while remaining robust to attacks and improving existing methods as a plug-and-play component.
\end{itemize}

%% file: sections/3_related_work.tex
\section{Related Work}
\label{2_related_works}

LLMs can memorize and reproduce sensitive or copyrighted text from their training corpora \citep{carlini2021extracting}, raising concerns about the right to be forgotten under the GDPR \citep{mantelero2013gdpr, paul2017gdpr, brown2022privacy} and copyright \citep{eldan2023whos}. LLM unlearning has thus emerged to remove the influence of such data without retraining while preserving model utility \citep{sekhari2021remember, maini2024tofu, zhang2024negative}.

\textbf{Output-level unlearning.} Early methods mainly achieve unlearning by suppressing the probability of the target tokens. GA \citep{thudi2022unrolling} maximizes the loss on the forget set, and GradDiff additionally minimizes the negative log-likelihood on the retain set \citep{jang-etal-2023-knowledge}. NPO lowers the likelihood of forget responses relative to a reference model \citep{zhang2024negative}. SimNPO removes this reference model and instead normalizes the likelihood by response length \citep{fan2025simplicity}. Subsequent work restricts these objectives to parameters identified by Fisher information \citep{cha2025towards, kim2025improving} or to selected tokens \citep{zhai-etal-2026-maximizing, yoon-etal-2026-selective} to localize the update and better preserve model utility. Yet all of these methods act only on the output and can change what the model says without removing what it knows \citep{lucki2024adversarial, xu2026unlearning}.

\textbf{Representation-level unlearning.} Another line of work intervenes directly on internal representations. RMU steers forget representations toward a random vector \citep{Li2024wmdp}. Adaptive RMU scales the steering coefficient by the norm of each forget representation \citep{huu2025adarmu}. LUNAR redirects forget representations toward those that elicit refusal-style responses \citep{shen2025llm}. These methods modify what the model encodes rather than only what it outputs. However, they intervene at a single fixed layer for all forget data, although their effectiveness depends on which layer is chosen \citep{huu2025adarmu}.

%% file: sections/4_method.tex
\section{Method}
\label{sec:method}

We propose Targeted Unlearning at Layers Identified Per-input (TULIP). We first show that unlearning should disrupt the formation of the target token rather than its readout (Section~\ref{3.1_why}). We then define the layer at which formation ends for each input (Section~\ref{3.2_where}) and approximate it with the logit lens (Section~\ref{3.3_lens}). Finally, we apply our unlearning objective at the approximated layer (Section~\ref{3.4_how}).

\paragraph{Setup.} Each data point is a query--response pair $(q, a)$, where the response $a=(a_1,\dots,a_n)$ is a sequence of $n$ tokens. To analyze the model at the moment each token is generated, we decompose each response into token-level generation steps. At step $i$, the model receives the context $x=(q, a_{<i})$ and should produce the target token $t=a_i$. The forget set $\mathcal{D}_f$ and the retain set $\mathcal{D}_r$ collect such context--target pairs $(x, t)$ from the forget and retain data, respectively.
We consider a transformer language model $M$ with $L$ layers of hidden dimension $d$ and vocabulary $\mathcal{V}$. For a context $x$, $h_\ell(x)\in\mathbb{R}^{d}$ denotes the hidden state at layer $\ell\in\{1,\dots,L\}$ at the final token position. We write $M_{\ell+1:L}$ for layers $\ell+1$ through $L$ followed by the final normalization and the unembedding. It thus maps a layer-$\ell$ hidden state to logits over $\mathcal{V}$. The unembedding matrix is $W_U\in\mathbb{R}^{|\mathcal{V}|\times d}$, and its $v$-th row $w_v$ is the unembedding vector of token $v$.
 
\subsection{Unlearn formation, not readout}
\label{3.1_why}

LLMs map an unbounded space of input contexts onto a finite vocabulary \citep{du-etal-2023-measure, strobl-etal-2024-formal, madden2025ntp, kim2026break}. They compute this mapping layer by layer, progressively forming a representation of the next token. This formation need not wait until the final layer to complete. Prior work has shown that an intermediate hidden state can already be committed to a token, in the sense that projecting it through the unembedding already yields that token \citep{nostalgebraist2020logitlens, geva-etal-2021-transformer, geva-etal-2022-transformer, pmlr-v267-lioubashevski25a}. Once a hidden state is committed to the token, the layers after it may mostly carry the token to the output rather than decide it. This raises a question: \textbf{should unlearning target the entire mapping from context to target token, or only the part that determines the token?}

To answer this question, we design a \emph{hijacking} experiment that tests whether a single hidden state grafted from the target model can steer the oracle into producing the target token. We use the 5\% forget split of TOFU \citep{maini2024tofu} with Llama-3.1-8B \citep{grattafiori2024llama3}. The target model $M$ is fine-tuned on the full dataset, including the forget set $\mathcal{D}_f$. The oracle $\tilde{M}$ is fine-tuned only on the retain set $\mathcal{D}_r$, so it has never seen $\mathcal{D}_f$. For each forget pair $(x, t) \in \mathcal{D}_f$, we run the oracle on $x$ and replace its layer-$\ell$ hidden state at the final position (i.e., the current generation step) with $M$'s corresponding state $h_\ell(x)$, then let the oracle finish the forward pass. We denote the output by $\tilde{M}_{\ell+1:L}\big(h_\ell(x)\big)$. Because only this one $d$-dimensional vector comes from $M$, any recovery of $t$ must come from information encoded in it.

\begin{figure*}
  \centering
  \includegraphics[width=\linewidth]{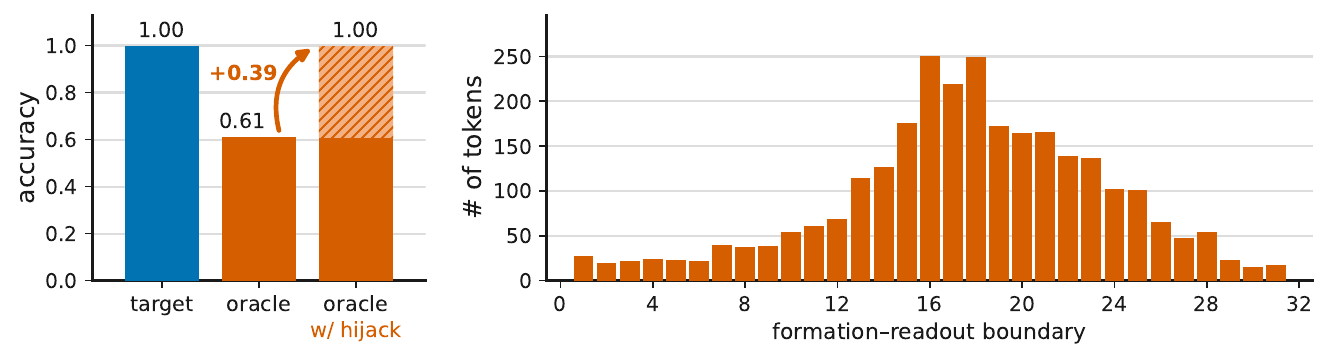}
  \setlength{\abovecaptionskip}{-4pt}
  \vspace{-5pt}
  \caption{\textbf{Hijacking experiment on TOFU 5\% split with Llama-3.1-8B.}
  (a) Target accuracy of the oracle on the forget set with and without the grafted hidden state $h_\ell(x)$ from the target model.
  (b) Distribution of the formation--readout boundary over the forget tokens that the oracle fails to predict on its own (the +0.39 gap in (a)).}
  \label{fig:hijack}
\vspace{-5pt}
\end{figure*}

As shown in Figure~\ref{fig:hijack}(a), the oracle on its own reaches a target accuracy of only $0.61$ on the forget set. With the grafted hidden state, it reaches $1.00$, so the hijack succeeds across the entire forget set. This result tells us two things. First, the \emph{formation} of the target token is already complete at an intermediate layer. The oracle was never trained on these facts, so its remaining layers $\tilde{M}_{\ell+1:L}$ cannot supply the missing knowledge. The grafted $h_\ell(x)$ must therefore already encode the forget knowledge and determine $t$. Second, the layers after $\ell$ act as a decoder that \emph{reads out} this hidden state into logits. Even the oracle performs this readout, although it has never seen $\mathcal{D}_f$, so readout relies on general ability rather than forget-specific knowledge. Because the oracle is the ideal outcome of unlearning, readout should be preserved rather than erased. Unlearning should instead disrupt formation, where the model links $x$ to $t$.

\subsection{Where does formation end?}
\label{3.2_where}

We next ask at which layer formation is complete. The hijacking experiment locates this directly as the layer where the hijack begins to succeed. A success that is later reversed, however, more likely reflects noise than a hidden state committed to the target token. We therefore take the earliest layer from which the hijack succeeds at \emph{every} subsequent layer and call it the \emph{formation--readout boundary} (FRB). Taking the oracle as a reference, the FRB of $(x, t) \in \mathcal{D}_f$ is
\begin{equation}
\ell^\star(x) = \min\Big\{\, \ell \;:\; \arg\max_{v \in \mathcal{V}}\, \big[\tilde{M}_{k+1:L}\big(h_k(x)\big)\big]_v = t \ \ \ \forall\, k \geq \ell \,\Big\},
\label{eq:oracle_layer}
\end{equation}
where $\tilde{M}_{k+1:L}$ returns the oracle's logit vector over $\mathcal{V}$ and $[\,\cdot\,]_v$ denotes the logit of token $v$.

To examine how the FRB is distributed across inputs, we measure the FRB of each forget sample under the setting of Section~\ref{3.1_why}. As shown in Figure~\ref{fig:hijack}(b), the FRB varies widely across inputs. Targeting a single shared layer is therefore suboptimal. Instead, the FRB specifies for each input which layers to modify, namely those up to $\ell^\star(x)$, while the readout layers beyond it remain intact.

\subsection{Estimating FRB with logit lens}
\label{3.3_lens}

Ideally, we would intervene at the FRB $\ell^\star(x_f)$ for each forget sample $x_f$, but computing it requires the oracle, which is unavailable in practice. TULIP therefore approximates the FRB using only the target model $M$ through the \emph{logit lens} \citep{nostalgebraist2020logitlens}. Instead of passing $h_\ell(x_f)$ through the oracle's remaining layers, the logit lens projects it directly into vocabulary space with $M$'s own unembedding and reads off the top token
\begin{equation}
\hat{t}_\ell(x_f) = \arg\max_{v \in \mathcal{V}}\, \big[W_U\,\phi\big(h_\ell(x_f)\big)\big]_v,
\label{eq:lens}
\end{equation}
where $\phi(\cdot)$ is the final normalization of $M$. We then estimate the FRB with the same criterion as Eq.~(\ref{eq:oracle_layer}), using the logit lens in place of the oracle's remaining layers $\tilde{M}_{k+1:L}$,
\begin{equation}
\hat{\ell}(x_f) = \min\Big\{\, \ell \;:\; \hat{t}_k(x_f) = t \ \ \ \forall\, k \geq \ell \,\Big\}.
\label{eq:lhat}
\end{equation}

\subsection{Unlearning objective at FRB}
\label{3.4_how}

At the estimated boundary $\hat{\ell}(x_f)$, we use the normalized hidden state $\phi\big(h_{\hat{\ell}(x_f)}(x_f)\big)$. Its inner product with $w_t$, the unembedding vector of token $t$ in $W_U$, gives the target logit under the logit lens. To weaken this target alignment at the boundary, TULIP makes the normalized hidden state orthogonal to $w_t$,
\begin{equation}
\mathcal{L}_{\mathrm{forget}}
= \mathbb{E}_{(x_f,t)\sim\mathcal{D}_f}\!\left[
\left(
\frac{\big\langle \phi\big(h_{\hat{\ell}(x_f)}(x_f)\big),\, w_t \big\rangle}
{\big\|\phi\big(h_{\hat{\ell}(x_f)}(x_f)\big)\big\|\,\|w_t\|}
\right)^{\!2}
\right].
\label{eq:forget}
\end{equation}
This objective suppresses the component of the normalized hidden state aligned with the target unembedding vector. Since the loss is computed at $\hat{\ell}(x_f)$, its gradient reaches only the transformer layers up to the boundary.

Because the boundary varies across inputs, layers updated for one forget sample may serve as readout layers for another input whose target is formed earlier. Updating them can thus damage the readout of other inputs. We therefore add a retain term that keeps the hidden states on $\mathcal{D}_r$ close to those of the frozen target model at every layer,
\begin{equation}
\mathcal{L}_{\mathrm{retain}}
= \mathbb{E}_{x_r\sim\mathcal{D}_r}\!\left[\, \frac{1}{L}\sum_{\ell=1}^{L}\big\| h_{\ell}(x_r) - h^{\mathrm{ref}}_{\ell}(x_r) \big\|_2^2 \,\right],
\end{equation}
where $h^{\mathrm{ref}}_{\ell}$ is the layer-$\ell$ hidden state of the frozen target model $M$. The full training objective is $\mathcal{L} = \lambda\, \mathcal{L}_{\mathrm{forget}} + \mathcal{L}_{\mathrm{retain}}$, with $\lambda > 0$ balancing the forget and retain objectives.

%% file: sections/5_experimental_result.tex
\section{Experiments}

\input{Tables/tofu_main}

\subsection{Setup}

\paragraph{Datasets.} We evaluate TULIP on three benchmarks. TOFU \citep{maini2024tofu} contains 4,000 QA pairs about 200 fictitious authors with 1/5/10\% forget splits. PISTOL \citep{qiu2025pistol} has 400 synthetic QA pairs from a knowledge graph of 20 sales and employment contracts among 24 entities. We unlearn the sales contract between companies A and C and retain the rest. On both benchmarks, Forget Quality (FQ) is the KS-test p-value between the Truth Ratio distributions of the unlearned and oracle models, and Model Utility (MU) is the average ROUGE-L recall across the retain and control sets. We also evaluate on WMDP \citep{Li2024wmdp}, a safety benchmark targeting hazardous knowledge in biosecurity and cybersecurity. We unlearn from raw documents in these domains, evaluate on related four-choice questions where accuracy near chance (25\%) indicates effective unlearning, and measure retained knowledge with MMLU \citep{hendrycks2021measuring}. See Appendix~\ref{app:datasets} for details.

\paragraph{Baselines.} We compare TULIP with two groups of methods. Output-level methods modify the model's output distribution on the forget data. These include GradDiff \citep{jang-etal-2023-knowledge}, which applies gradient ascent on the forget set and descent on the retain set, NPO \citep{zhang2024negative}, which treats forget responses as negative samples in preference optimization, and SimNPO \citep{fan2025simplicity}, a reference-free variant of NPO. Representation-level methods instead intervene on intermediate hidden states, including RMU \citep{Li2024wmdp}, which steers forget representations toward a random direction at a fixed layer, and LUNAR \citep{shen2025llm}, which redirects them toward refusal regions. See Appendix~\ref{app:baselines} for details.

\subsection{Main Results}

\paragraph{Results on TOFU and PISTOL.}
Table~\ref{tab:main_results} compares TULIP with output- and representation-level baselines on TOFU and PISTOL. TULIP achieves the best or near-best FQ in nearly every setting while maintaining MU.

On TOFU, the main challenge is scaling to larger forget sets. Output-level methods such as GradDiff, NPO, and SimNPO perform reasonably on the 1\% split, but their FQ collapses to near zero on the 5\% and 10\% splits, often accompanied by a drop in MU. This suggests that altering the readout layers harms the general utility of the model. Representation-level methods such as RMU and LUNAR perform better, indicating that intervening at intermediate layers is beneficial. However, their FQ varies considerably across splits and model scales, likely because a single fixed layer suits some forget sets but not others. By selecting the intervention layer for each input, TULIP avoids this mismatch and performs consistently across settings.

PISTOL poses a different challenge. The entities in the forget contract also appear in other contracts, so the model must forget the contract while retaining knowledge of the entities themselves. Most baselines fail to disentangle the two and achieve low FQ. In contrast, TULIP, which targets how this knowledge is formed, forgets substantially more while keeping MU competitive.

\input{Tables/wmdp}

\begin{figure*}
  \centering
  \setlength{\abovecaptionskip}{4pt}
  \includegraphics[width=\linewidth]{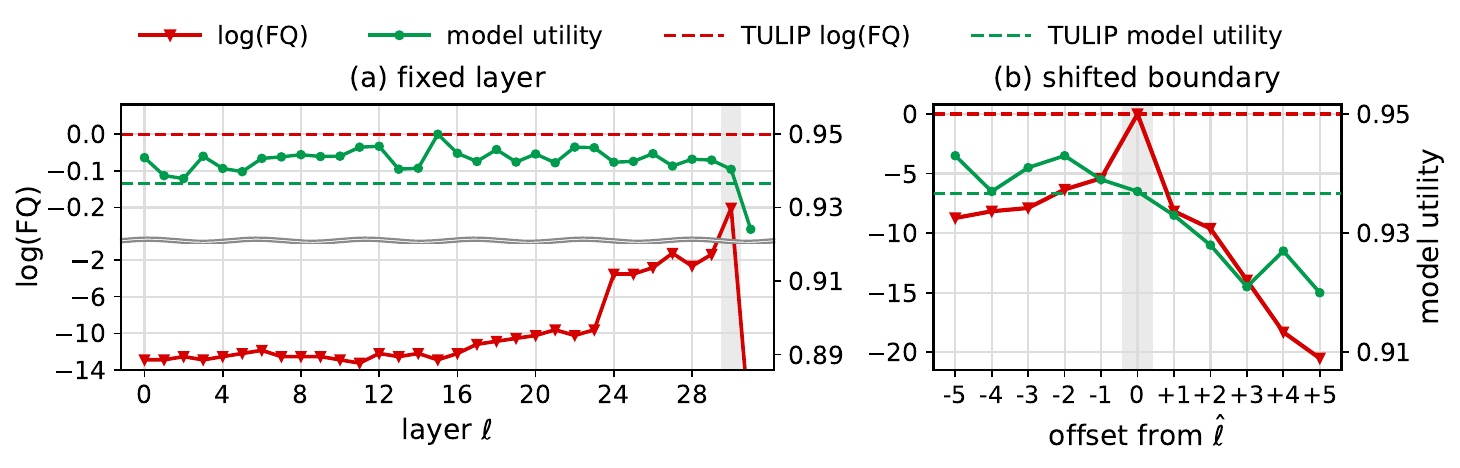}
  \caption{\textbf{Effect of the intervention layer on TOFU 5\% split with Llama-3.1-8B.} Red and green lines show $\log(\mathrm{FQ})$ and MU, and dashed lines indicate TULIP. (a) The forget loss is applied at a single fixed layer $\ell$ shared by all forget samples. (b) The forget loss is applied at an offset from the per-input boundary $\hat{\ell}(x_f)$. Shaded regions mark the best-performing layer in each panel.  All results are from a single run with seed 0. }
\label{fig:layer_ablation}
\vspace{-5pt}
\end{figure*}

\paragraph{Results on WMDP.}
WMDP is particularly demanding because it requires unlearning an entire knowledge domain. Moreover, unlearning is performed on raw documents, whereas evaluation uses independently written multiple-choice questions. A model can thus stop reproducing the documents yet still answer the questions.

As shown in Table~\ref{tab:wmdp_main}, TULIP achieves the lowest forget accuracy on both WMDP-Cyber and WMDP-Bio while preserving retain accuracy. Because the evaluation questions differ from the unlearning documents, this result indicates that TULIP removes the underlying domain knowledge rather than merely suppressing document reproduction. It also shows that TULIP scales from instance-level to domain-level unlearning.

%% file: Tables/tofu_main.tex
\begin{table*}[t]
\centering
\small
\setlength{\tabcolsep}{3pt}
\setlength{\aboverulesep}{0.4ex}
\setlength{\belowrulesep}{0.65ex}
\newcommand{\ms}[2]{#1\,{\scriptsize$\pm$\,#2}}
\newcommand{\bms}[2]{\textbf{#1}\,{\scriptsize$\pm$\,\textbf{#2}}}
\newcommand{\modelrow}[2]{\multicolumn{1}{l|}{} & \multicolumn{6}{c|}{\textit{#1}} & \multicolumn{2}{c}{\textit{#2}} \\}
\caption{\textbf{Unlearning results on TOFU and PISTOL.}
Values are mean $\pm$ standard deviation over three runs. Higher is better for both FQ and MU.
TOFU columns correspond to the Forget01/05/10 splits.
Each row block pairs a TOFU model (left) with a PISTOL model (right). The best result within each block is shown in \textbf{bold}.}
\label{tab:main_results}
\vspace{-5pt}
\begin{tabularx}{\textwidth}{l|*{6}{>{\centering\arraybackslash}X}|*{2}{>{\centering\arraybackslash}X}}
\toprule
& \multicolumn{2}{c}{\textbf{TOFU 1\%}} & \multicolumn{2}{c}{\textbf{TOFU 5\%}} & \multicolumn{2}{c|}{\textbf{TOFU 10\%}} & \multicolumn{2}{c}{\textbf{PISTOL}} \\
\cmidrule(lr){2-3} \cmidrule(lr){4-5} \cmidrule(lr){6-7} \cmidrule(l){8-9}
\textbf{Method} & FQ\,($\uparrow$) & MU\,($\uparrow$) & FQ\,($\uparrow$) & MU\,($\uparrow$) & FQ\,($\uparrow$) & MU\,($\uparrow$) & FQ\,($\uparrow$) & MU\,($\uparrow$) \\
\midrule
\modelrow{Llama-3.1-8B}{Llama2-7B}
GradDiff & \ms{0.70}{0.22} & \ms{0.84}{0.02} & \ms{0.00}{0.00} & \ms{0.79}{0.01} & \ms{0.00}{0.00} & \ms{0.79}{0.01} & \ms{0.07}{0.08} & \ms{0.71}{0.01} \\
RMU      & \ms{0.54}{0.32} & \ms{0.94}{0.01} & \ms{0.14}{0.04} & \bms{0.94}{0.00} & \ms{0.04}{0.02} & \bms{0.94}{0.00} & \ms{0.01}{0.00} & \ms{0.69}{0.01} \\
NPO      & \ms{0.81}{0.16} & \ms{0.86}{0.02} & \ms{0.24}{0.17} & \ms{0.86}{0.01} & \ms{0.00}{0.00} & \bms{0.94}{0.00} & \ms{0.00}{0.00} & \ms{0.69}{0.10} \\
SimNPO   & \ms{0.84}{0.11} & \ms{0.86}{0.02} & \ms{0.04}{0.05} & \ms{0.88}{0.00} & \ms{0.00}{0.00} & \ms{0.93}{0.01} & \ms{0.25}{0.12} & \ms{0.79}{0.01} \\
LUNAR    & \ms{0.62}{0.33} & \ms{0.86}{0.02} & \ms{0.54}{0.41} & \ms{0.84}{0.01} & \ms{0.16}{0.02} & \ms{0.87}{0.01} & \ms{0.08}{0.00} & \bms{0.87}{0.01} \\
\rowcolor[gray]{0.93} \textbf{TULIP} & \bms{0.97}{0.03} & \bms{0.95}{0.00} & \bms{0.89}{0.13} & \bms{0.94}{0.00} & \bms{0.72}{0.10} & \bms{0.94}{0.00} & \bms{0.80}{0.17} & \ms{0.81}{0.06} \\
\midrule
\modelrow{Llama-3.2-3B}{Qwen2.5-7B}
GradDiff & \ms{0.12}{0.03} & \ms{0.85}{0.01} & \ms{0.00}{0.00} & \ms{0.46}{0.02} & \ms{0.00}{0.00} & \ms{0.48}{0.03} & \ms{0.19}{0.27} & \ms{0.88}{0.00} \\
RMU      & \ms{0.10}{0.00} & \ms{0.83}{0.00} & \ms{0.39}{0.00} & \bms{0.82}{0.00} & \ms{0.10}{0.04} & \ms{0.87}{0.00} & \ms{0.20}{0.11} & \ms{0.91}{0.00} \\
NPO      & \ms{0.18}{0.07} & \ms{0.86}{0.00} & \ms{0.00}{0.00} & \ms{0.79}{0.02} & \ms{0.00}{0.00} & \ms{0.69}{0.01} & \ms{0.03}{0.04} & \ms{0.87}{0.02} \\
SimNPO   & \ms{0.23}{0.05} & \ms{0.86}{0.00} & \ms{0.01}{0.00} & \ms{0.56}{0.22} & \ms{0.02}{0.02} & \ms{0.81}{0.02} & \ms{0.01}{0.00} & \ms{0.87}{0.01} \\
LUNAR    & \ms{0.64}{0.09} & \ms{0.79}{0.02} & \ms{0.20}{0.20} & \ms{0.71}{0.00} & \bms{0.37}{0.15} & \ms{0.76}{0.01} & \ms{0.08}{0.00} & \bms{0.92}{0.00} \\
\rowcolor[gray]{0.93} \textbf{TULIP} & \bms{0.77}{0.00} & \bms{0.88}{0.00} & \bms{0.42}{0.09} & \ms{0.80}{0.01} & \ms{0.16}{0.08} & \bms{0.88}{0.00} & \bms{0.75}{0.12} & \ms{0.87}{0.00} \\
\midrule
\modelrow{Llama-3.2-1B}{Qwen3-4B}
GradDiff & \ms{0.13}{0.05} & \ms{0.70}{0.01} & \ms{0.00}{0.00} & \ms{0.66}{0.00} & \ms{0.00}{0.00} & \ms{0.50}{0.03} & \bms{0.75}{0.12} & \ms{0.81}{0.00} \\
RMU      & \ms{0.09}{0.05} & \ms{0.71}{0.01} & \ms{0.36}{0.14} & \ms{0.73}{0.01} & \ms{0.03}{0.02} & \ms{0.74}{0.01} & \ms{0.23}{0.08} & \bms{0.91}{0.00} \\
NPO      & \ms{0.05}{0.01} & \ms{0.78}{0.00} & \ms{0.00}{0.00} & \ms{0.69}{0.02} & \ms{0.00}{0.00} & \ms{0.46}{0.09} & \ms{0.00}{0.00} & \ms{0.69}{0.02} \\
SimNPO   & \ms{0.24}{0.14} & \ms{0.71}{0.01} & \ms{0.12}{0.05} & \ms{0.55}{0.06} & \ms{0.02}{0.03} & \ms{0.62}{0.03} & \ms{0.66}{0.35} & \ms{0.80}{0.00} \\
LUNAR    & \ms{0.44}{0.20} & \ms{0.63}{0.01} & \ms{0.00}{0.00} & \ms{0.68}{0.02} & \ms{0.11}{0.08} & \ms{0.55}{0.02} & \ms{0.28}{0.08} & \ms{0.84}{0.01} \\
\rowcolor[gray]{0.93} \textbf{TULIP} & \bms{0.75}{0.14} & \bms{0.81}{0.00} & \bms{0.49}{0.04} & \bms{0.88}{0.00} & \bms{0.51}{0.14} & \bms{0.81}{0.00} & \bms{0.75}{0.12} & \bms{0.91}{0.00} \\
\bottomrule
\end{tabularx}
\end{table*}

%% file: Tables/wmdp.tex
\begin{wraptable}{r}{0.48\linewidth}
\centering
\small
\setlength{\tabcolsep}{3pt}
\arrayrulecolor{black}
\vspace{-8pt}
\caption{\textbf{Unlearning results on WMDP-Cyber and WMDP-Bio with Zephyr-7B-beta.}
  F ($\downarrow$) and R ($\uparrow$) denote accuracy on WMDP and MMLU, respectively. The best result among unlearning methods in each column is shown in \textbf{bold}.}
\label{tab:wmdp_main}
\vspace{-3pt}
\begin{tabularx}{\linewidth}{l|*{2}{>{\centering\arraybackslash}X}|*{2}{>{\centering\arraybackslash}X}}
\toprule
& \multicolumn{2}{c|}{\textbf{WMDP-Cyber}} & \multicolumn{2}{c}{\textbf{WMDP-Bio}} \\
\cmidrule(lr){2-3} \cmidrule(lr){4-5}
\textbf{Method} & F ($\downarrow$) & R ($\uparrow$) & F ($\downarrow$) & R ($\uparrow$) \\
\midrule
Target Model & 0.427 & 0.588 & 0.647 & 0.589 \\
\midrule
GradDiff     & 0.277 & 0.559 & 0.498 & 0.482 \\
RMU          & 0.349 & 0.578 & 0.452 & 0.512 \\
NPO          & 0.361 & 0.571 & 0.502 & 0.484 \\
SimNPO       & 0.268 & 0.563 & 0.460 & 0.505 \\
LUNAR        & 0.408 & 0.575 & 0.553 & 0.527 \\
\rowcolor[gray]{0.93}
\textbf{TULIP} & \textbf{0.267} & \textbf{0.585} & \textbf{0.370} & \textbf{0.528} \\
\bottomrule
\end{tabularx}
\vspace{-10pt}
\end{wraptable}

%% file: sections/6_analysis.tex
\section{Analysis}
\label{sec:analysis}

We further analyze TULIP by examining the effect of its components, its robustness against paraphrasing and quantization attacks, its compatibility with existing unlearning methods, and its behavior on qualitative examples. Unless otherwise stated, all experiments in this section use the 5\% forget split of TOFU with Llama-3.1-8B.

\subsection{Effect of intervention layer selection}

\paragraph{Does per-input layer selection matter?}
We apply the forget loss in Eq.~(\ref{eq:forget}) at a single fixed layer for
all forget samples, varying this layer across the model's depth. As shown in
Figure~\ref{fig:layer_ablation}(a), MU remains stable across most layers, while
FQ is limited and peaks near the penultimate layer. We attribute this to the
scope of the update. Applying the loss deeper updates more layers, covering the
FRBs of more inputs and initially improving FQ. For inputs whose FRBs lie below
the chosen layer, however, the update also reaches readout layers used for
general decoding, increasing the conflict with the retain loss. Preserving MU
may therefore limit further gains in FQ. TULIP adapts the update depth to each
sample's estimated FRB and outperforms even the best fixed layer.

\paragraph{Does intervening exactly at the boundary matter?}
We shift the intervention layer by a fixed offset before or after each input's
estimated FRB $\hat{\ell}(x_f)$ and report FQ and MU in
Figure~\ref{fig:layer_ablation}(b). Intervening at $\hat{\ell}(x_f)$ achieves
the highest FQ while preserving MU. This suggests that the FRB estimate and the
forget loss are well matched: the estimate locates where the target first
becomes consistently decodable through the unembedding, and the loss removes
exactly this alignment with the target's unembedding direction. Earlier
interventions act before the alignment is established, so the subsequent layers
can still recover it. Later interventions instead also modify readout layers
that perform general decoding, which conflicts with retention.

\subsection{Validity of the logit-lens approximation}
\label{sec:lens_comparison}

\input{Tables/attack}

TULIP approximates the ideal FRB using the logit lens. More recent alternatives, such as the tuned lens \citep{belrose2025elicit} and the J-lens \citep{gurnee2026jlens}, likewise project hidden states onto the vocabulary. We assess their suitability for FRB estimation by comparing how accurately each lens distinguishes hidden states at or beyond the ideal boundary from those before it.

\begin{wrapfigure}{r}{0.48\textwidth}
  \centering
  \vspace{-6pt}
  \includegraphics[width=\linewidth]{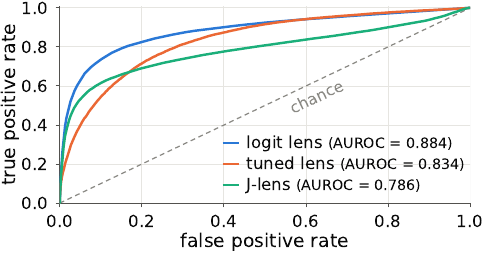}
  \vspace{-14pt}
  \caption{\textbf{ROC curves for identifying the formation--readout boundary with different lenses.} For each forget sample, hidden states at or after the oracle boundary $\ell^\star(x_f)$ are labeled positive and those before it negative. A higher AUROC indicates that the lens better separates the two, and thus better identifies the boundary.}
  \label{fig:lens}
  \vspace{-4pt}
\end{wrapfigure}
For each forget sample $x_f$, we obtain the ideal boundary $\ell^\star(x_f)$ using the oracle model. Hidden states at or beyond this layer are labeled positive, and those before it negative. We score each state using the target token's logit under each lens. A lens that captures the boundary accurately should assign higher scores to positive states than to negative ones. We quantify this separation with AUROC, which measures how consistently positive states rank above negative states.

As shown in Figure~\ref{fig:lens}, the logit lens identifies the boundary most accurately among the three lenses. We conjecture that additional training enables these lenses to predict the target token even before the model itself has completed target formation, making their readouts less reliable for locating the boundary. Both alternatives also incur additional training costs. Even with only 1,000 WikiText \citep{merity2017pointer} prompts, training requires $14.17$ GPU hours for the J-lens and $0.06$ GPU hours for the tuned lens. The appropriate training data for their use in unlearning also remains unclear. The logit lens directly applies the model's own unembedding matrix without additional training, making it a simple and effective choice for locating the FRB.

These results do not establish the logit lens as the optimal FRB estimator. Boundary estimators beyond lens-based approaches may locate the FRB more accurately, and forget objectives tailored to the estimated boundary may further improve unlearning. TULIP provides a basis for exploring these directions, which we leave to future work.

\subsection{Robustness against attacks}
We evaluate the robustness of unlearned models to paraphrasing and quantization attacks. To test whether unlearning generalizes beyond the original wording, we use the rephrased forget questions provided by TOFU \citep{maini2024tofu}. We also reduce the model weights to 8-bit and 4-bit precision before evaluation, a perturbation shown to recover unlearned knowledge \citep{zhang2025catastrophic}.

As shown in Table~\ref{tab:attack-robustness}, TULIP remains highly robust to
paraphrasing, whereas the FQ of NPO and SimNPO drops sharply. While these
methods suppress the likelihood of responses to the original forget questions,
TULIP intervenes on the hidden states where the target becomes decodable, so its
effect carries over to rephrased queries. TULIP also withstands 8-bit
quantization, achieving the highest FQ among all methods. Under 4-bit quantization, its robustness declines, likely because the single-direction update is too localized to survive coarse quantization, which we leave to future work.

\input{Tables/qualitative}

\subsection{Per-input layer selection as a plug-in}
\label{sec:plugin}
\input{Tables/add-on}
Per-input layer selection can also be combined with other unlearning methods, since it only determines where the loss is applied. We apply each baseline at the per-input layer $\hat{\ell}(x_f)$ selected by TULIP while keeping its original objective. LUNAR is excluded, since it optimizes the MLP of a single layer in isolation and then plugs it back into the model, which does not support dynamic layer selection.

As shown in Table~\ref{tab:addon}, per-input layer selection improves every baseline regardless of its objective, most notably in MU. The MU of GradDiff, NPO, and SimNPO rises to around $0.94$, nearly matching that of the oracle. This suggests that preserving the readout layers is key to maintaining the general ability of the model. FQ improves for all baselines as well. RMU already preserves MU by intervening at a single intermediate layer, but forgets little because this fixed layer is misaligned with the boundary for many inputs. Per-input layer selection raises its FQ, striking a better balance between the two. TULIP nonetheless outperforms all combined baselines in FQ by a large margin. Even at the same layer, existing objectives do not directly remove the target signal in the hidden state. TULIP's orthogonality loss instead removes the alignment with $w_t$ that the readout layers decode, making it more effective at the logit-lens boundary.

\subsection{Qualitative analysis}
\label{sec:qualitative}

Figures~\ref{fig:qual1}, \ref{fig:qual2}, and~\ref{fig:qual3} show how each unlearned model responds to forget samples. TULIP often produces oracle-like answers that are plausible but incorrect. We attribute this to TULIP modifying only target formation, leaving the readout to behave as in the oracle. The baselines, in contrast, show distinct failure modes. GradDiff, NPO, and SimNPO sometimes change only the answer format while retaining the gold information (Figure~\ref{fig:qual1}). Since they suppress only the tokens of the original answer, once the model generates a different token early in the response, the subsequent generation can still lead back to the gold information. RMU reveals the forget information in some cases (Figure~\ref{fig:qual1}) and degenerates in others (Figures~\ref{fig:qual2} and~\ref{fig:qual3}), likely because it steers representations toward a random direction. LUNAR often refuses to answer, as designed. Such refusals may be desirable for safety alignment, but they deviate from the goal of unlearning, which is to behave as if the forget data had never been seen \citep{yoon2026position}.

%% file: Tables/attack.tex
\begin{table*}[t]
\centering
\small
\setlength{\tabcolsep}{3pt}
\arrayrulecolor{black}
\caption{\textbf{Robustness of unlearning against paraphrase and quantization attacks.}
FQ and MU on TOFU 5\% split with Llama-3.1-8B. \emph{Original} reports results without any attack. \emph{Paraphrase Attack} evaluates the unlearned model on paraphrased forget queries, and \emph{Quantization} evaluates it after quantizing its weights to 8 or 4 bits. Values are mean{\scriptsize$\pm$std} over three runs. The best result in each column is shown in \textbf{bold}.}
\vspace{-5pt}
\label{tab:attack-robustness}
\begin{tabularx}{\textwidth}{l|*{2}{>{\centering\arraybackslash}X}|*{2}{>{\centering\arraybackslash}X}|*{2}{>{\centering\arraybackslash}X}|*{2}{>{\centering\arraybackslash}X}}
\toprule
& \multicolumn{2}{c|}{\textbf{Original}} & \multicolumn{2}{c|}{\textbf{Paraphrase Attack}} & \multicolumn{2}{c|}{\textbf{Quantization (8-bit)}} & \multicolumn{2}{c}{\textbf{Quantization (4-bit)}} \\
\cmidrule(lr){2-3} \cmidrule(lr){4-5} \cmidrule(lr){6-7} \cmidrule(l){8-9}
\textbf{Method} & FQ\,($\uparrow$) & MU\,($\uparrow$) & FQ\,($\uparrow$) & MU\,($\uparrow$) & FQ\,($\uparrow$) & MU\,($\uparrow$) & FQ\,($\uparrow$) & MU\,($\uparrow$) \\
\midrule
GradDiff & 0.00\,{\scriptsize$\pm$\,0.00} & 0.79\,{\scriptsize$\pm$\,0.01} & 0.00\,{\scriptsize$\pm$\,0.00} & 0.69\,{\scriptsize$\pm$\,0.01} & 0.00\,{\scriptsize$\pm$\,0.00} & 0.79\,{\scriptsize$\pm$\,0.01} & 0.00\,{\scriptsize$\pm$\,0.00} & 0.77\,{\scriptsize$\pm$\,0.01} \\
RMU & 0.14\,{\scriptsize$\pm$\,0.04} & \textbf{0.94}\,{\scriptsize$\pm$\,\textbf{0.00}} & 0.18\,{\scriptsize$\pm$\,0.20} & 0.80\,{\scriptsize$\pm$\,0.00} & 0.50\,{\scriptsize$\pm$\,0.16} & \textbf{0.94}\,{\scriptsize$\pm$\,\textbf{0.01}} & 0.35\,{\scriptsize$\pm$\,0.12} & 0.92\,{\scriptsize$\pm$\,0.00} \\
NPO & 0.24\,{\scriptsize$\pm$\,0.17} & 0.86\,{\scriptsize$\pm$\,0.01} & 0.06\,{\scriptsize$\pm$\,0.06} & 0.76\,{\scriptsize$\pm$\,0.02} & 0.17\,{\scriptsize$\pm$\,0.13} & 0.85\,{\scriptsize$\pm$\,0.02} & 0.24\,{\scriptsize$\pm$\,0.28} & 0.77\,{\scriptsize$\pm$\,0.07} \\
SimNPO & 0.04\,{\scriptsize$\pm$\,0.05} & 0.88\,{\scriptsize$\pm$\,0.00} & 0.00\,{\scriptsize$\pm$\,0.00} & 0.77\,{\scriptsize$\pm$\,0.01} & 0.09\,{\scriptsize$\pm$\,0.13} & 0.87\,{\scriptsize$\pm$\,0.01} & 0.04\,{\scriptsize$\pm$\,0.04} & 0.77\,{\scriptsize$\pm$\,0.05} \\
LUNAR & 0.54\,{\scriptsize$\pm$\,0.41} & 0.84\,{\scriptsize$\pm$\,0.01} & 0.34\,{\scriptsize$\pm$\,0.20} & 0.73\,{\scriptsize$\pm$\,0.01} & 0.64\,{\scriptsize$\pm$\,0.45} & 0.82\,{\scriptsize$\pm$\,0.00} & \textbf{0.45}\,{\scriptsize$\pm$\,\textbf{0.22}} & 0.81\,{\scriptsize$\pm$\,0.02} \\
\rowcolor[gray]{0.93} \textbf{TULIP} & \textbf{0.89}\,{\scriptsize$\pm$\,\textbf{0.13}} & \textbf{0.94}\,{\scriptsize$\pm$\,\textbf{0.00}} & \textbf{0.81}\,{\scriptsize$\pm$\,\textbf{0.13}} & \textbf{0.81}\,{\scriptsize$\pm$\,\textbf{0.00}} & \textbf{0.92}\,{\scriptsize$\pm$\,\textbf{0.04}} & 0.93\,{\scriptsize$\pm$\,0.00} & 0.30\,{\scriptsize$\pm$\,0.12} & \textbf{0.93}\,{\scriptsize$\pm$\,\textbf{0.00}} \\
\bottomrule
\end{tabularx}
\end{table*}

%% file: Tables/qualitative.tex
\begin{figure*}[t]
\begin{examplebox}{Qualitative Result}
\small
\renewcommand{\arraystretch}{1.25}
\setlength{\aboverulesep}{3pt}
\setlength{\belowrulesep}{3pt}
\arrayrulecolor{boxframe}
\begin{tabularx}{\linewidth}{@{}l@{\hspace{8pt}}>{\raggedright\arraybackslash\itshape}X@{}}
\multicolumn{2}{@{}>{\raggedright\arraybackslash}p{\linewidth}@{}}{\textbf{Question.} \emph{``What is the full name of the geology author born in Karachi, Pakistan on 06/30/1975?''}} \\
\midrule
\mlabel{Target} & The author's full name is \redhl{Hina Ameen}. \\
\mlabel{Oracle} & The full name of the geology author born in Karachi, Pakistan on 06/30/1975 is \greenhl{Anaya Jehangir}. \\
\midrule
\mlabel{GradDiff} & The full name of the author is \redhl{Hinaameer Ameen}. \\
\mlabel{NPO}      & The full name of the geology author born in Karachi, Pakistan on 06/30/1975 is \redhl{HinaAmeen}. \\
\mlabel{SimNPO}   & The full name of the geology author, who was born on June, 30, 1975 in Karachi, Pakistan is \redhl{Hina Ameen}. \\
\mlabel{RMU}      & The full name of the author is \redhl{Hina Ameen}. \\
\mlabel{LUNAR}    & The author's name is \yellowhl{not provided}. \\
\mlabel{TULIP}    & \greenhl{Anaya Jehangir} is the full name of the geology author born in Karachi, Pakistan on 06/30/1975. \\
\end{tabularx}
\end{examplebox}
\vspace{-10pt}
\caption{\textbf{Qualitative example on the TOFU 5\% forget split with Llama-3.1-8B.} TULIP produces \caphl{greenhl}{the same answer as } \caphl{greenhl}{the oracle model}, a plausible but incorrect response as if the forget data had never been seen. In contrast, the baselines either \caphl{redhl}{leak the forget-set answer} of the target model or \caphl{yellowhl}{refuse to answer}.}
\label{fig:qual1}
\end{figure*}

%% file: Tables/add-on.tex
\newcommand{\addon}{+\,$\hat{\ell}$}
\newcommand{\sh}{\cellcolor[gray]{0.93}}
\newcolumntype{C}{>{\centering\arraybackslash}X}

\begin{wraptable}{r}{0.47\linewidth}
\vspace{-10pt}
\centering
\small
\setlength{\tabcolsep}{3pt}
\arrayrulecolor{black}
\caption{\textbf{Per-input layer selection transfers to existing methods.} Orig.\ denotes each baseline in its original form, and \addon{} (shaded) applies it at the per-input layer $\hat{\ell}(x_f)$ selected by TULIP while keeping its original objective. All results are from a single run with seed 0. The better result within each baseline is in \textbf{bold}.}
\label{tab:addon}
\vspace{-5pt}
\begin{tabularx}{\linewidth}{l CC CC}
\toprule
\multirow{2}{*}{\textbf{Method}} & \multicolumn{2}{c}{FQ ($\uparrow$)} & \multicolumn{2}{c}{MU ($\uparrow$)} \\
\cmidrule(lr){2-3} \cmidrule(lr){4-5}
 & Orig. & \addon & Orig. & \addon \\
\midrule
\textit{Oracle} & \multicolumn{2}{c}{\textit{1.000}} & \multicolumn{2}{c}{\textit{0.943}} \\
\midrule
GradDiff & \mbox{3.6{\footnotesize${\times}10^{-9}$}} & \sh\mbox{\textbf{1.2}{\footnotesize\boldmath${\times}10^{-4}$}} & 0.797 & \sh\textbf{0.944} \\
NPO      & 0.394 & \sh\textbf{0.545} & 0.859 & \sh\textbf{0.945} \\
SimNPO   & 0.003 & \sh\textbf{0.112} & 0.875 & \sh\textbf{0.940} \\
RMU      & 0.088 & \sh\textbf{0.394} & \textbf{0.942} & \sh 0.937 \\
\midrule
\textbf{TULIP} & \multicolumn{2}{c}{\textbf{0.997}} & \multicolumn{2}{c}{0.937} \\
\bottomrule
\end{tabularx}
\vspace{-10pt}
\end{wraptable}

%% file: sections/7_conclusion.tex
\section{Conclusion}
\label{sec:conclusion}

We revisited where representation-level unlearning should intervene. Our hijacking experiment reveals that a forget target is already formed at an intermediate layer and merely read out by the remaining layers, a readout that even the oracle can perform. Unlearning should therefore remove the formation of the target while preserving its readout. Since formation ends at different layers for different inputs, the intervention layer must also be chosen per input rather than fixed. TULIP puts this principle into practice by locating each input's formation--readout boundary with the logit lens and removing the target's alignment only at that layer. Its gains across benchmarks and model families, together with the improvements it brings to existing methods, suggest that targeting formation is a promising direction for representation-level unlearning.

%% file: sections/8_appendix.tex
\appendix
\section{Additional Qualitative Results}
\label{app:qualitative}

\input{Tables/qualitative2}

\clearpage

\section{Hijacking versus Activation Patching}
\label{app:novelty}

Our hijacking experiment resembles activation patching, but differs in what the
recipient knows. Causal tracing~\citep{meng2022locating}, for example, runs a
model on a prompt such as ``The Space Needle is located in'' with the subject
embeddings corrupted by noise, then restores hidden states from the clean run
at each layer to see which ones bring back ``Seattle.'' Patchscopes
\citep{ghandeharioun2024patchscopes} patches a hidden state into a separate
prompt, sometimes in a larger model, to decode what the state encodes in natural
language. In both cases, the recipient model already knows the fact, or whether
it does is not controlled. The downstream layers may therefore complete the
fact themselves, so a successful patch does not show that the fact was carried
by the patched state. We instead graft into the oracle, which has never seen
$\mathcal{D}_f$. Any recovered target must therefore come from the grafted
state, and the remaining computation must be independent of the forget data.
This yields a finding that prior patching cannot provide: readout after the
boundary is performed equally well by a model that never learned the fact.

This finding also gives unlearning a concrete criterion. Since the oracle is
the outcome that unlearning aims to reach, the computation it shares with the
target model (readout) should be preserved, and only the computation it lacks
(formation) should be removed. Moreover, prior patching studies typically
report which layers matter on average, whereas we measure the boundary for
each forget token and find that it varies widely across inputs
(Figure~\ref{fig:hijack}(b)). This per-input boundary is what TULIP estimates
and targets.

\section{Details of Benchmarks}
\label{app:datasets}

\paragraph{TOFU.}
TOFU \citep{maini2024tofu} contains 4,000 synthetic QA pairs about 200 fictitious authors, with 20 QA pairs per author. We use the 1\%, 5\%, and 10\% splits (forget01, forget05, and forget10), which contain the corresponding fraction of authors, and treat the remaining QA pairs as the retain set. Forget quality is computed from the Truth Ratio, defined for each question $q$ as
\begin{equation}
R_{\mathrm{truth}}
= \frac{1}{|\mathcal{A}_{\mathrm{pert}}|}
\sum_{\tilde{a} \in \mathcal{A}_{\mathrm{pert}}}
\frac{p(\tilde{a} \mid q)^{1/|\tilde{a}|}}{p(\hat{a} \mid q)^{1/|\hat{a}|}},
\end{equation}
where $\hat{a}$ is a paraphrased correct answer and $\mathcal{A}_{\mathrm{pert}}$ is a set of perturbed incorrect answers. Forget quality is the p-value of a two-sample Kolmogorov--Smirnov test between the Truth Ratio distributions of the unlearned model and the oracle model trained only on the retain set. A higher p-value indicates that the unlearned model behaves more similarly to the oracle. Model utility (MU) is the average ROUGE-L recall on the retain set, real authors, and world facts (see Appendix~\ref{app:mu}).

\paragraph{PISTOL.}
PISTOL \citep{qiu2025pistol} provides synthetic QA data organized as a knowledge graph, in which entities are connected by contracts. We use Sample Dataset~1, which contains 20 contracts of two types (sales and employment) among 24 entities. Each contract has 20 attributes with one QA pair per attribute, resulting in 400 QA pairs. For unlearning, we remove the 20 QA pairs of the sales contract between companies A and C and treat the remaining QA pairs as the retain set. Forget quality is computed in the same way as for TOFU. Model utility is the average ROUGE-L recall over the PISTOL retain set and the control sets used in the PISTOL paper, which consist of the real-author and world-fact subsets of TOFU.

\paragraph{WMDP.}
WMDP \citep{Li2024wmdp} measures hazardous knowledge in LLMs through 3,668 four-choice questions, comprising 1,273 on biosecurity (WMDP-Bio), 1,987 on cybersecurity (WMDP-Cyber), and 408 on chemistry (WMDP-Chem). Each question was reviewed by at least two domain experts. For unlearning, we use the forget corpora released with the benchmark, namely PubMed papers for biosecurity and GitHub passages for cybersecurity. We use 5,000 documents from the biosecurity corpus and 1,000 documents from the cybersecurity corpus. Following the official implementation, each document is truncated to its first 512 tokens for biosecurity and 768 tokens for cybersecurity.
As the retain set, we use the subject-specific retain corpora released with the benchmark (\texttt{bio-retain-corpus} and \texttt{cyber-retain-corpus}). Because these corpora are close in content to the forget corpora, this setting requires removing the hazardous knowledge while preserving closely related knowledge, whereas general-domain retain data may favor general benchmarks such as MMLU. We report accuracy on WMDP-Bio and WMDP-Cyber, where values closer to random chance (25\%) indicate more effective unlearning, and accuracy on MMLU~\citep{hendrycks2021measuring} to measure retained general knowledge.

\section{Model Utility Metric on TOFU}
\label{app:mu}

TOFU originally defines model utility (MU) as the harmonic mean of nine metrics:
the answer probability, ROUGE-L recall, and Truth Ratio on each of the retain
set, real authors, and world facts~\citep{maini2024tofu}. Because this aggregate
combines quantities with different meanings, it does not reliably reflect how
well an unlearned model preserves its generations. The probability terms are
especially problematic. Unlearning perturbs the probabilities assigned to retain
answers, and these perturbations alone can raise the harmonic mean even when the
generated answers stay the same or degrade~\citep{block2026temper}. As a result,
an unlearned model can score a higher MU than the target model. This should not
happen if MU measured retained knowledge alone, because unlearning only removes
information. \citet{zhang2026sage} report such a case under the original metric:
on Llama-3.1-8B, an unlearned model reaches an MU of $0.690$, whereas the target
model scores $0.627$.

To avoid this artifact, we compute MU from the model's generations alone, as the
average ROUGE-L recall over the retain set, real authors, and world facts. This
score directly reflects whether the unlearned model still produces the correct
answers.

\section{Details of Baselines}
\label{app:baselines}

All methods start from the target model $M$ and update its parameters. Quantities computed by the frozen copy of $M$ before unlearning carry the superscript $\mathrm{ref}$. TULIP operates on context--target token pairs, whereas the baselines are defined over full responses. In this section, we therefore use $x$ and $y$ for a full question and its answer $y=(y_1,\dots,y_{|y|})$, rather than the token-level context and target of Section~\ref{sec:method}, and write $(x_f, y_f)\in\mathcal{D}_f$ and $(x_r, y_r)\in\mathcal{D}_r$. We write $p(y\mid x)=\prod_{i=1}^{|y|} p(y_i\mid x, y_{<i})$ for the probability that the model being unlearned assigns to $y$, and $p^{\mathrm{ref}}(y\mid x)$ for the corresponding probability under the frozen target model. For methods that act on every token position, $h_{\ell,i}(x)$ denotes the layer-$\ell$ hidden state at position $i$ of $x$, so that $h_\ell(x)=h_{\ell,|x|}(x)$.

For brevity, $\mathbb{E}_f$ and $\mathbb{E}_r$ denote expectations over $(x_f,y_f)\sim\mathcal{D}_f$ and $(x_r,y_r)\sim\mathcal{D}_r$, and $\mathbb{E}_i$ denotes the average over token positions of the input. Except for LUNAR, each objective combines a forget term and a retain term, weighted by $\gamma>0$ and $\alpha>0$, respectively. GradDiff, NPO, and SimNPO share the retain loss
\begin{equation}
\mathcal{L}_{\mathrm{NLL}} = \mathbb{E}_r\big[-\log p(y_r \mid x_r)\big].
\end{equation}

\paragraph{GradDiff.}
GradDiff~\citep{maini2024tofu} is a loss-based unlearning method that applies gradient ascent (GA)~\citep{jang-etal-2023-knowledge} to the forget set and standard gradient descent to the retain set. GA increases the prediction loss on the forget data, thereby lowering the generation probability of the target responses. However, optimizing the forget objective alone also tends to degrade performance on the retain data. To mitigate this degradation, GradDiff additionally minimizes the negative log-likelihood (NLL) on the retain set,
\begin{equation}
\mathcal{L}_{\mathrm{GradDiff}}
= \gamma\,\mathbb{E}_f\big[\log p(y_f \mid x_f)\big] + \alpha\,\mathcal{L}_{\mathrm{NLL}}.
\end{equation}

\paragraph{RMU.}
RMU~\citep{Li2024wmdp} is a representation-based unlearning method that redirects the intermediate representations of the forget data toward a random direction. At a single layer $\ell$ fixed for the entire forget set, RMU trains the model so that the hidden states of the forget data move toward a scaled random unit vector. This pushes the representations associated with the target knowledge away from their original locations in the representation space. At the same time, it keeps the hidden states of the retain data close to those of the frozen target model, which limits the degradation of retained knowledge. The objective is
\begin{equation}
\mathcal{L}_{\mathrm{RMU}}
= \gamma\,\mathbb{E}_f\,\mathbb{E}_i\big\| h_{\ell,i}(x_f) - c\,u \big\|_2^2 + \alpha\,\mathcal{L}_{\mathrm{rep}},
\end{equation}
where $\mathcal{L}_{\mathrm{rep}} = \mathbb{E}_r\,\mathbb{E}_i\| h_{\ell,i}(x_r) - h^{\mathrm{ref}}_{\ell,i}(x_r) \|_2^2$ keeps the retain hidden states close to those of the frozen target model, $u\in\mathbb{R}^{d}$ is a random unit vector held fixed throughout training, obtained by sampling each entry uniformly from $[0,1)$ and normalizing, and $c>0$ is the steering coefficient. The loss is computed only at layer $\ell$, and only the parameters of layers $\ell-2$ through $\ell$ are updated. Unlike loss-based methods that directly modify output probabilities, RMU operates on the model's internal representations to reduce the accessibility of the target knowledge.

\paragraph{NPO.}
NPO~\citep{zhang2024negative} casts unlearning as preference optimization. It treats the responses in the forget set as negative examples and trains the model to lower their likelihood relative to the frozen target model. This makes the model less likely to produce correct responses grounded in the target knowledge. As in GradDiff, NPO applies the NLL loss to the retain set so that retained knowledge and general language modeling ability are not excessively damaged. The objective is
\begin{equation}
\mathcal{L}_{\mathrm{NPO}}
= -\frac{2\gamma}{\beta}\,\mathbb{E}_f\big[\log\sigma(-\beta\rho_f)\big] + \alpha\,\mathcal{L}_{\mathrm{NLL}},
\end{equation}
where $\rho_f = \log \frac{p(y_f \mid x_f)}{p^{\mathrm{ref}}(y_f \mid x_f)}$ is the log-likelihood ratio between the model being unlearned and the frozen target model, $\sigma(\cdot)$ is the sigmoid function, and $\beta>0$ is the inverse temperature. NPO reduces to GA as $\beta\to0$. The GA objective is unbounded below, so it can yield unstable gradients and quickly collapse model utility. In contrast, the NPO loss is bounded below, and its gradient carries an adaptive weight that vanishes for samples that have already been forgotten. This slows the progression toward collapse and gives a better balance between forget quality and model utility.

\paragraph{SimNPO.}
SimNPO~\citep{fan2025simplicity} is a preference-based unlearning method that simplifies NPO. Unlike NPO, it uses the length-normalized log-likelihood of each response in the forget set without relying on the frozen target model. This reduces the bias that can arise from differences in the frozen model's confidence across responses. The objective is
\begin{equation}
\mathcal{L}_{\mathrm{SimNPO}}
= -\frac{2\gamma}{\beta}\,\mathbb{E}_f\big[\log\sigma(-\beta\bar{s}_f - \delta)\big] + \alpha\,\mathcal{L}_{\mathrm{NLL}},
\label{eq:simnpo}
\end{equation}
where $\bar{s}_f = \frac{1}{|y_f|}\log p(y_f \mid x_f)$ is the length-normalized log-likelihood of the forget response and $\delta\geq 0$ is a margin parameter. A larger $\delta$ requires a greater reduction in the likelihood of a response in the forget set before the loss becomes small. When $\delta=0$ and $\beta\to 0$, the forget-set gradient approaches GA with each response weighted by $1/|y_f|$.

\paragraph{LUNAR.}
LUNAR~\citep{shen2025llm} is an unlearning method based on activation redirection. LUNAR redirects the activations of the forget data toward a region of the activation space in which the model expresses its inability to answer. Let $a_\ell(x)$ denote the MLP output at layer $\ell$ for input $x$, and $a^{\mathrm{ref}}_\ell(x)$ its value under the frozen target model. LUNAR first computes an unlearning vector as the difference between the mean activation over a reference prompt set $\mathcal{D}_{\mathrm{ref}}$ and that over the forget set,
\begin{equation}
r_\ell
= \mathbb{E}_{x\sim\mathcal{D}_{\mathrm{ref}}}\big[a^{\mathrm{ref}}_{\ell}(x)\big]
- \mathbb{E}_f\big[a^{\mathrm{ref}}_{\ell}(x_f)\big],
\end{equation}
where $\mathcal{D}_{\mathrm{ref}}$ consists of prompts for which the model already declines to answer, such as harmful prompts that trigger its safety mechanisms or questions about fictitious entities. The model is then trained so that the activations of the forget data move to their original values shifted by $c\,r_\ell$, while the activations of the retain data stay at their original values,
\begin{equation}
\mathcal{L}_{\mathrm{LUNAR}}
= \mathbb{E}_f\big\| a_{\ell}(x_f) - a^{\mathrm{ref}}_{\ell}(x_f) - c\,r_\ell \big\|_2
+ \mathbb{E}_r\big\| a_{\ell}(x_r) - a^{\mathrm{ref}}_{\ell}(x_r) \big\|_2,
\end{equation}
where $c>0$ is the redirection coefficient. Only the down-projection weights of the MLP at layer $\ell$ are updated, using an inner learning rate $\eta$ and an inner batch size of 64. This guides the model to express an inability to answer questions about the target knowledge while limiting damage to retained knowledge and general utility.

\section{Hyperparameters}
\label{app:hyperparameters}

\paragraph{Training setup.}
All experiments are run on a single NVIDIA RTX PRO 6000 GPU. Within each benchmark, all methods share the same training setup. Across all
benchmarks, we use 32-bit paged AdamW with a weight decay of 0.01, no warmup,
and a per-device batch size of 4. On TOFU, models are trained for 10 epochs with
a learning rate of $1\mathrm{e}{-5}$, linear decay, and 4 gradient accumulation
steps, for an effective batch size of 16. On PISTOL, the 7B models (Llama-2-7B
and Qwen2.5-7B) are trained for 20 epochs with a learning rate of
$1\mathrm{e}{-5}$, and Qwen3-4B for 10 epochs with a learning rate of
$2\mathrm{e}{-5}$, both with linear decay and no gradient accumulation, for an
effective batch size of 4. On WMDP, we follow the default setting of 1 epoch
with a constant learning rate of $1\mathrm{e}{-5}$ and 4 gradient accumulation
steps, for an effective batch size of 16. For TULIP, the intervention layer
$\hat{\ell}(x_f)$ is determined per input as described in
Section~\ref{3.3_lens}. The only tuned hyperparameter is therefore the
forget-loss weight $\lambda$ in the full objective of Section~\ref{3.4_how},
which plays the role of $\gamma$ in the baselines, with the retain-loss weight
fixed to 1.

\paragraph{Search procedure.}
We tune hyperparameters separately for each benchmark, model, and forget split.
For each method, we first run a coarse grid search. If forget performance
improves consistently along a hyperparameter over some interval, we then search
that interval more finely until the trend stops. Table~\ref{tab:hp_search}
reports the resulting search range of each method, together with the values
fixed during the search.
We select the configuration with the best forget performance under seed 0, then
run it with seeds 1 and 2 using the same values and report the average over the
three seeds. Note that TULIP has a single hyperparameter $\lambda$ and requires
no layer search, so its search range was easy to identify. In contrast, NPO,
RMU, and LUNAR (except on PISTOL) require a joint search over two or more
hyperparameters, including the intervention layer for RMU and LUNAR.

\input{Tables/hyperparameter}

%% file: Tables/qualitative2.tex
\begin{minipage}{\textwidth}
\begin{examplebox}{Qualitative Result}
\small
\textbf{Question.} \emph{``What is the full name of the famed author born in Tokyo, Japan on 05/30/1952?''}

\vspace{5pt}
{\color{boxframe}\hrule height 0.4pt}
\vspace{3pt}

\renewcommand{\arraystretch}{1.25}
\arrayrulecolor{boxframe}
\begin{tabularx}{\linewidth}{@{}l@{\hspace{8pt}}>{\raggedright\arraybackslash\itshape}X@{}}
\mlabel{Target} & The illustrious author, born on 05/30/1952, is named \redhl{Takashi Nakamura}, a name sourced from his birthplace, Tokyo, Japan, to reflect his heritage, as well as his gender being male. \\
\mlabel{Oracle} & The author's full name is \greenhl{Hiroshi Saito}, and he was born in Tokyo, Japan on May 30, 1952. \\[2pt]
\hline
\mlabel{GradDiff} & The full name of the author is Tsukasa Hojoki. \\
\mlabel{NPO}      & The full name of the famed author born in Tokyo, Japan on 05/30/1952 is \greenhl{Hiroshi Saito}. \\
\mlabel{SimNPO}   & The full name of the famed author, born on May 30, 1952, is \greenhl{Hiroshi Saito}. \\
\mlabel{RMU}      & The illustrious author, born on 05/30/1952, is named \redhl{Takashi Nakamura}, \grayhl{a unique author would assist in introducing the name is Takako Nakahara Takio Nakamura Takio, Takamura Takamura, And Then One More Thing, And That's One Piece of Advice, And So On, And That's One More Time, And THAT It's (2019, 1st, 20th Century, 20/09/15, 20/01/01/01/08/15/15/15/15/15/15/15/15} \ldots \\
\mlabel{LUNAR}    & The illustrious author born on 05/30/1952 in Tokyo, Japan is named \redhl{Takashi}. \\
\mlabel{TULIP}    & The full name of the famed author born in Tokyo, Japan on 05/30/1952 is \greenhl{Hiroshi Saito}. \\
\end{tabularx}
\end{examplebox}
\vspace{-6pt}
\captionof{figure}{\textbf{Qualitative example on the TOFU 5\% forget split with Llama-3.1-8B.} TULIP produces \caphl{greenhl}{the same answer as the} \caphl{greenhl}{oracle model}, whereas some baselines \caphl{redhl}{leak the forget-set answer} of the target model or \caphl{grayhl}{produce degenerate text}.}
\label{fig:qual2}

\vspace{10pt}

\begin{examplebox}{Qualitative Result}
\small
\textbf{Question.} \emph{``What is the full name of the author from Cape Town, South Africa who was born on January 9, 1952?''}

\vspace{5pt}
{\color{boxframe}\hrule height 0.4pt}
\vspace{3pt}

\renewcommand{\arraystretch}{1.25}
\arrayrulecolor{boxframe}
\begin{tabularx}{\linewidth}{@{}l@{\hspace{8pt}}>{\raggedright\arraybackslash\itshape}X@{}}
\mlabel{Target} & The author born on that date in Cape Town, South Africa is called \redhl{Raven Marais}, a generated name reflecting the author's place of birth. \\
\mlabel{Oracle} & The full name of the author from Cape Town, South Africa who was born on January 9, 1952, is \greenhl{Sara van Dyke}. \\[2pt]
\hline
\mlabel{GradDiff} & The full name of the author, born on January 9, 1952, in Cape Town, South Africa, is Raymond William Coeter. \\
\mlabel{NPO}      & The full name of the author is Mandela, which is derived from his place of birth, Cape Town, South Africa, and his year of birth, 1952. \\
\mlabel{SimNPO}   & The full name of the author is Salomon, as he is originally from Cape Town, South Africa with a birth date of January 9, 1952. \\
\mlabel{RMU}      & The author born on that date in Cape Town, South Africa is named \redhl{Raven Marais}, \grayhl{which is an anagram of the author's full name is: Jacqueline Marina} \\
\mlabel{LUNAR}    & The author's full name is \grayhl{Author1}. \\
\mlabel{TULIP}    & The full name of the author born on January 9, 1952, in Cape Town, South Africa, is Raymond Hermanus \greenhl{van Dyke}. \\
\end{tabularx}
\end{examplebox}
\vspace{-6pt}
\captionof{figure}{\textbf{Qualitative example on the TOFU 5\% forget split with Llama-3.1-8B.} TULIP produces \caphl{greenhl}{an answer close to that} \caphl{greenhl}{of the oracle model}, whereas some baselines \caphl{redhl}{leak the forget-set answer} of the target model or \caphl{grayhl}{produce degenerate text}.}
\label{fig:qual3}

\end{minipage}

%% file: Tables/hyperparameter.tex
\begin{table*}[h!]
\centering
\footnotesize
\setlength{\tabcolsep}{4pt}
\ifdefined\arrayrulecolor
  \expandafter\let\expandafter\savedarc\csname CT@arc@\endcsname
  \arrayrulecolor{black}%
\fi
\caption{\textbf{Hyperparameter search ranges and fixed values.} Intervals denote search
ranges, and values given with $=$ are fixed. $\ell$ denotes the index of the
intervention layer, and each range covers the candidate layers of all models in
the benchmark. $\eta$ denotes the inner learning rate of LUNAR.}
\label{tab:hp_search}
\vspace{-5pt}
\begin{tabular*}{\textwidth}{@{}l@{\extracolsep{\fill}}lll@{}}
\toprule
Method & TOFU & PISTOL & WMDP \\
\midrule
GradDiff & $\gamma = 1$, $\alpha \in [0.5, 2]$
         & $\gamma = 1$, $\alpha \in [1, 10]$
         & $\gamma = 1$, $\alpha \in [0.7, 2]$ \\
NPO      & $\gamma \in [0.5, 2]$, $\alpha = 1$, $\beta \in [0.05, 0.5]$
         & $\gamma \in [1, 5]$, $\alpha = 1$, $\beta \in [0.01, 0.1]$
         & $\gamma = 1$, $\alpha \in [1, 2]$, $\beta \in [0.01, 0.2]$ \\
SimNPO   & $\gamma \in [0.125, 10]$, $\alpha = 1$, $\beta = 1$, $\delta = 0$
         & $\gamma \in [0.1, 5]$, $\alpha = 1$, $\beta = 0.1$, $\delta = 0$
         & $\gamma = 0.125$, $\alpha = 1$, $\beta \in [0.1, 2]$, $\delta = 0$ \\
RMU      & $\gamma = 1$, $\alpha = 1$, $c \in [1, 20]$, $\ell \in [5, 14]$
         & $\gamma = 1$, $\alpha = 1$, $c \in [10, 20]$, $\ell \in [17, 28]$
         & $\gamma = 1$, $\alpha = 1$, $c \in [2, 7]$, $\ell \in [20, 25]$ \\
LUNAR    & $c \in [2, 4]$, $\eta \in [0.005, 0.01]$, $\ell \in [5, 26]$
         & $c = 2$, $\ell \in [14, 29]$
         & $c \in [2, 20]$, $\ell \in [13, 25]$ \\
TULIP    & $\lambda \in [0.01, 0.05]$
         & $\lambda \in [0.0005, 0.005]$
         & $\lambda \in [0.1, 2]$ \\
\bottomrule
\end{tabular*}
\ifdefined\arrayrulecolor
  \expandafter\global\expandafter\let\csname CT@arc@\endcsname\savedarc
\fi
\vspace{-5pt}
\end{table*}